\documentclass[conference]{IEEEtran}

\IEEEoverridecommandlockouts
\usepackage{cite}
\usepackage{amsmath,amssymb,amsfonts}
\usepackage{hyperref}
\usepackage{float}
\usepackage{algorithm,algcompatible}

\algnewcommand\INPUT{\item[\textbf{Input:}]}%
\algnewcommand\OUTPUT{\item[\textbf{Output:}]}%

\usepackage{graphicx}
\usepackage{textcomp}
\usepackage{xcolor}
\def\BibTeX{{\rm B\kern-.05em{\sc i\kern-.025em b}\kern-.08em
    T\kern-.1667em\lower.7ex\hbox{E}\kern-.125emX}}
    
\usepackage{hyperref}
\hypersetup{
    colorlinks,
    linkcolor={black},
    urlcolor={blue!80!black}
}    
\usepackage{tikz}
\usepackage{xcolor}
\usepackage{pgfplots}
\usetikzlibrary{spy}
\pgfplotsset{compat=1.18} 
\begin{document}

\setlength{\columnsep}{0.24in}
\addtolength{\topmargin}{+0.047cm}

\thispagestyle{empty}
\pagestyle{empty}
\title{Equivariant Imaging for Self-supervised Hyperspectral Image Inpainting\\

}

\author{\IEEEauthorblockN{Shuo Li}
\IEEEauthorblockA{\textit{School of Engineering} \\
\textit{University of Edinburgh}\\
s1809498@ed.ac.uk}
\and
\IEEEauthorblockN{Mike Davies}
\IEEEauthorblockA{\textit{School of Engineering} \\
\textit{University of Edinburgh}\\
mike.davies@ed.ac.uk}
\and
\IEEEauthorblockN{Mehrdad Yaghoobi}
\IEEEauthorblockA{\textit{School of Engineering} \\
\textit{University of Edinburgh}\\
m.yaghoobi-vaighan@ed.ac.uk}
}

\maketitle

\begin{abstract}
Hyperspectral imaging (HSI) is a key technology for earth observation, surveillance, medical imaging and diagnostics, astronomy and space exploration. The conventional technology for HSI in remote sensing applications is based on the push-broom scanning approach in which the camera records the spectral image of a stripe of the scene at a time, while the image is generated by the aggregation of measurements through time. In real-world airborne and spaceborne HSI instruments, some empty stripes would appear at certain locations, because platforms do not always maintain a constant programmed attitude, or have access to accurate digital elevation maps (DEM), and the travelling track is not necessarily aligned with the hyperspectral cameras at all times. This makes the enhancement of the acquired HS images from incomplete or corrupted observations an essential task. We introduce a novel HSI inpainting algorithm here, called Hyperspectral Equivariant Imaging (Hyper-EI). Hyper-EI is a self-supervised learning-based method which does not require training on extensive datasets or access to a pre-trained model. Experimental results show that the proposed method achieves state-of-the-art inpainting performance compared to the existing methods.
\end{abstract}


\section{Introduction}
Due to the influence of instrumental errors, imperfect navigation and atmospheric changes, practical HSI images can suffer from missing (lines of) pixels, which make the use of measured data cubes challenging. As an example, the imaging system failures may lead to striped missing areas and cloud occlusion, resulting in irregular-shaped missing areas in HSIs. These issues can severely impact down-link tasks such as HSI classification, un-mixing and object detection. HSI inpainting involves restoring missing or corrupted data in acquired HS images. In contrast to RGB images, inpainting of HSIs requires filling in a complex vector that contains extensive spectral information, rather than just a single pixel value. The additional complexity makes the already challenging inpainting task even more difficult. \\
In the past decade, deep learning has revolutionized the ﬁeld of hyperspectral image processing, by learning and exploiting the intrinsic statistics of HS images that are critical for inpainting. In \cite{HSI-IPNet}, the authors proposed an attention-based Generative Adversarial Network (GAN), demonstrating that the missing pixels can be predicted through training on a large HSI dataset. Motivated by \cite{HSI-IPNet}, the authors in \cite{HSI_IP_GAN} modified the network architectures to capture the local detailed texture and improved the channel attention modules to leverage more effective frequency components. In \cite{GLCSA-Net} the attention modules are adopted in combination with a constraint to regularize the reconstruction considering from both the local and global perspectives. Despite the promising results, these end-to-end training solutions have several drawbacks: (1) They require training the networks on some extensive datasets, which can be cumbersome for data-hungry HSI applications, (2) They fail at recovering the sharp edges between the missing regions and non-missing regions, the inpainted areas tend to be blurry, and (3) These methods cannot be easily adapted to different sensor-specific challenges or different image resolutions, formats and domains, meaning they may need to be retrained or fine-tuned for different shapes of masks or different types of HS data. \\
As an alternative to fully-supervised HSI inpainting, the self-supervised method has recently gained significant attention for its data efficiency and generalization ability. 
In a groundbreaking study by \cite{DIP}, it was found that the structure of a generative network is sufficient to capture plenty of low-level image statistics before any learning. Their method has been extended to the HSI settings, called deep hyperspectral prior (DHP), and has been successfully applied to various HSI inverse tasks such as super-resolution, denoising and inpainting \cite{DHP}. The subsequent works \cite{DeepRED,pnp_dip,DIP-TV,R_DLRHyIn} further improved the performance of DHP by either combining it with off-the-shelf denoisers or other regularization priors. Recently, the authors in \cite{EI} demonstrated the feasibility of learning the inverse mapping directly from the incomplete measurement by adding a invariance assumption to the underlying signal set. This approach, known as equivariant imaging (EI), has achieved superior performance on various tasks such as: sparse-view CT image reconstruction \cite{chen2022robust}, super-resolution \cite{scanvic2023self}, and multispectral pansharpening \cite{wang2024perspective}.
In this paper, we proposed to solve HSI inpainting problems with EI, a self-supervised framework that can achieve high-quality inpainting through implicit access to multiple group operators.

\subsection{Contribution}
This paper aims to develop an effective HSI inpainting algorithm that does not need any external training data, i,e. self-supervised learning. Our contributions are as follows:
\begin{itemize}
\item Proposing a novel self-supervised algorithm, called Hyper-EI, which solves the challenging HSI inpainting tasks when the full spectral bands of a group of pixels may be missing, requiring only the corrupted HS data cube.
\item To exploit both spatial and spectral correlations of the HS images, we introduced a novel spatio-spectral attention architecture in this context to improve the inpainting performance.
\item Extensive experiments on real-world datasets demonstrate that the proposed algorithm outperforms existing self-supervised methods with superior inpainting performance on different datasets and masks, \textit{i.e.} generalisability and robustness. This observation is valuable in terms of challenging the common believe in the requirement of pre-trained models for high-quality HS inpainting. 
\end{itemize}

\section{Proposed Method}
\subsection{Hyper-EI Inpainting Algorithm}
The HSI inpainting task can be formulated as the reconstruction of the clean image $\boldsymbol{x}$ from the incomplete, or corrupted, measurement $\boldsymbol{y}$, in the presence of additive noise $n$ and masking operator $\boldsymbol{\rm M}$:
\begin{equation} \label{Formulation_HSI_Inpainting}
 \boldsymbol{y} = \boldsymbol{\rm M} \boldsymbol{x} +\boldsymbol{n}
\end{equation}
Recovering $\boldsymbol{x}$ from the measurement $\boldsymbol{y}$ is an ill-posed problem which requires prior knowledge of the underlying clean HS image $\boldsymbol{x}$ to be imposed during reconstruction. \\
The concept of EI was initially proposed in \cite{EI} where the authors showed that reconstruction only from the measurement data is possible if the group invariance operators can fully span the Null-space of the measurement operator. Invariance refers to the robustness against a certain group of transformations such that they provide a closed operation within the class of signal of interest. Taking natural images as an example, the visual content remains the same despite changes in the position/orientation of the image, hence, they are invariance to shifting and rotation.
\begin{figure}[h]
  \scalebox{0.96}{

\includegraphics[width=0.5\textwidth,height=0.3\textwidth,]{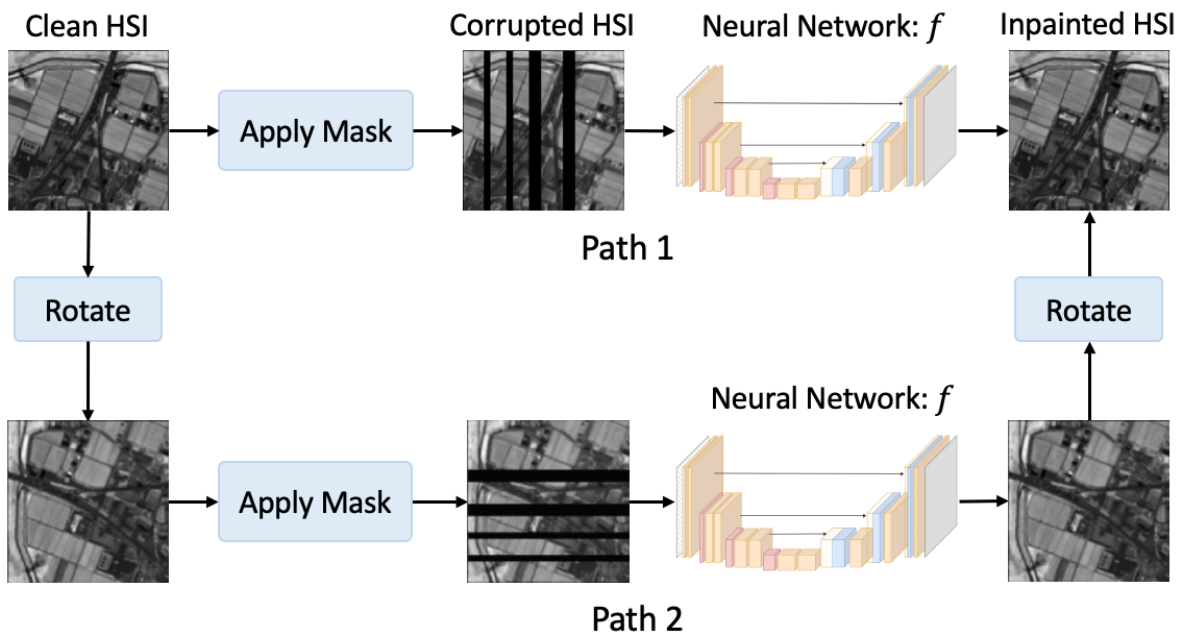}
 }
 \vspace{-0.5cm}
  \caption{Concept of equivariant imaging for HSI inpainting task problem. Path 1 and Path 2 represent the left-hand side and right-hand side of equation \eqref{EI_invariant}, respectively. The composition of the inpainting
operation with the reconstruction method $f$ (e,g. a neural network) should be equivariant to the rotation. Hence, both Path 1 and Path 2 yield identical results.}
  \label{EI_HSI_example}  
\end{figure} \\
A graphical example of EI is shown in Figure \ref{EI_HSI_example}, such a property brings new opportunities for solving image inverse problems as it can be enforced either implicitly as a prior, or explicitly through the architectural design of the neural networks \cite{chen2023_EI_related,tachella_EI_related}. Following the definition in \cite{EI}, we denote the group transformation as $\mathcal{G}\{\boldsymbol{g}_1,\boldsymbol{g}_2...\boldsymbol{g}_{\mathcal{|G|}}\}$, and a set of signals as $\mathcal{X} \subseteq \mathbb{R}^{q}$. Then, for all $ \boldsymbol{x} \in \mathcal{X}$ and $ \boldsymbol{g} \in \mathcal{G}$, the invariant property assumes that:
\begin{equation}\label{group_transformation_assumption}
     T_{g} \boldsymbol{x} \in \mathcal{X},  \quad T_{g}^{-1} \boldsymbol{x} \in \mathcal{X},  \quad
     T_{g}^{-1}(T_{g} \boldsymbol{x}) = \boldsymbol{x} 
\end{equation}
where $T_{g}, T_{g}^{-1} \in \mathbb{R}^{q \times q}$ are respectively the matrix form representation of group transformation $g$ and inverse of group operator transform , $T_{g}^{-1}T_{g} = T_{g}T_{g}^{-1} = I$ (e,g. $g$ can be rotation, shifting, or reflection and the inverse group transforms will be the same group operator transform but in the reverse direction. While we only consider linear transformation in the discrete signal domain, which can be presented using a matrix, the concept of group invariant transforms is more general and can be defined in the continuous domain and include nonlinear transforms. \\
\noindent In the image inpainting problem \eqref{Formulation_HSI_Inpainting}, the task is to learn the inverse mapping $f$ such that $f(\boldsymbol{y})=\boldsymbol{x} = T_{g}^{-1}T_{g} \boldsymbol{x}$. Combining the invariant assumption \eqref{group_transformation_assumption} and noiseless observation $\boldsymbol{y}$, we have:
\begin{equation}\label{invariant_inpainting_case}
\boldsymbol{y} = \mathbf{M} \boldsymbol{x} = \mathbf{M} T_{g}^{-1} T_{g} \boldsymbol{x} = (\mathbf{M} \circ T_{g}^{-1}) T_g\boldsymbol{x} = \Tilde{\mathbf{M}}\Tilde{\boldsymbol{x}}
\end{equation}
where $\tilde{\mathbf{x}} \in \mathcal{X}$ due to invariance property. For each measurement $\boldsymbol{y}$, the invariant property \eqref{invariant_inpainting_case} allows us to have access to a set of virtual operators $\Tilde{\mathbf{M}}$ which may have different null-spaces. In the context of Equivariant Imaging, the composition $f \circ \mathbf{M}$ should be invariant to the transformation, \textit{i.e.} commutativity of the two operators $f \circ \mathbf{M}$ and $T_g$, which gives the EI constraint:
\begin{equation}\label{EI_invariant}
(f \circ{\mathbf{M}})(T_{g}\boldsymbol{x}) = T_{g} (f \circ \mathbf{M})(\boldsymbol{x})
\end{equation}

\begin{figure}[h]  
 \vspace{-0.5cm}
\includegraphics[width=0.47 \textwidth]{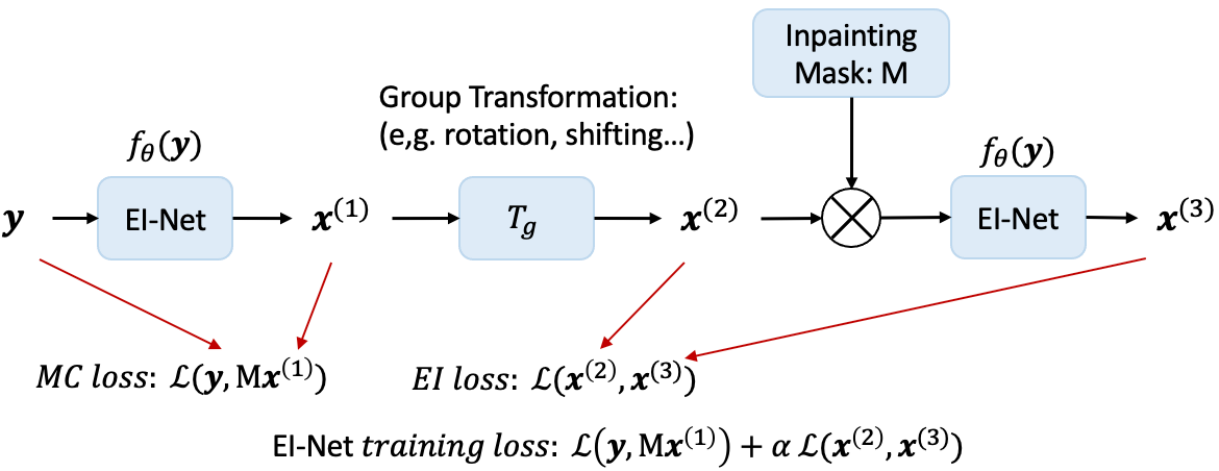}

  \caption{Training loss of the proposed Hyper-EI algorithm. the loss comprises the MC term and the EI regularization term.}
  \label{EI_training_loss}  
\end{figure}
\noindent The training process of the proposed Hyper-EI algorithm is shown in Figure \ref{EI_training_loss}, where $\boldsymbol{x^{(1)}}$ stands for the reconstructed image by EI-Net, $\boldsymbol{x^{(2)}}$ stands for the transformed $\boldsymbol{x^{(1)}}$, and finally, $\boldsymbol{x^{(3)}}$ stands for the re-estimation of $\boldsymbol{x^{(2)}}$. The training loss consists of two parts, namely, the Measurement 
\begin{figure*}
 \vspace{-0.58cm}
  \scalebox{1}{
  \hspace{0.2cm}
  \includegraphics[width=0.95\textwidth,height=0.72\textwidth,]{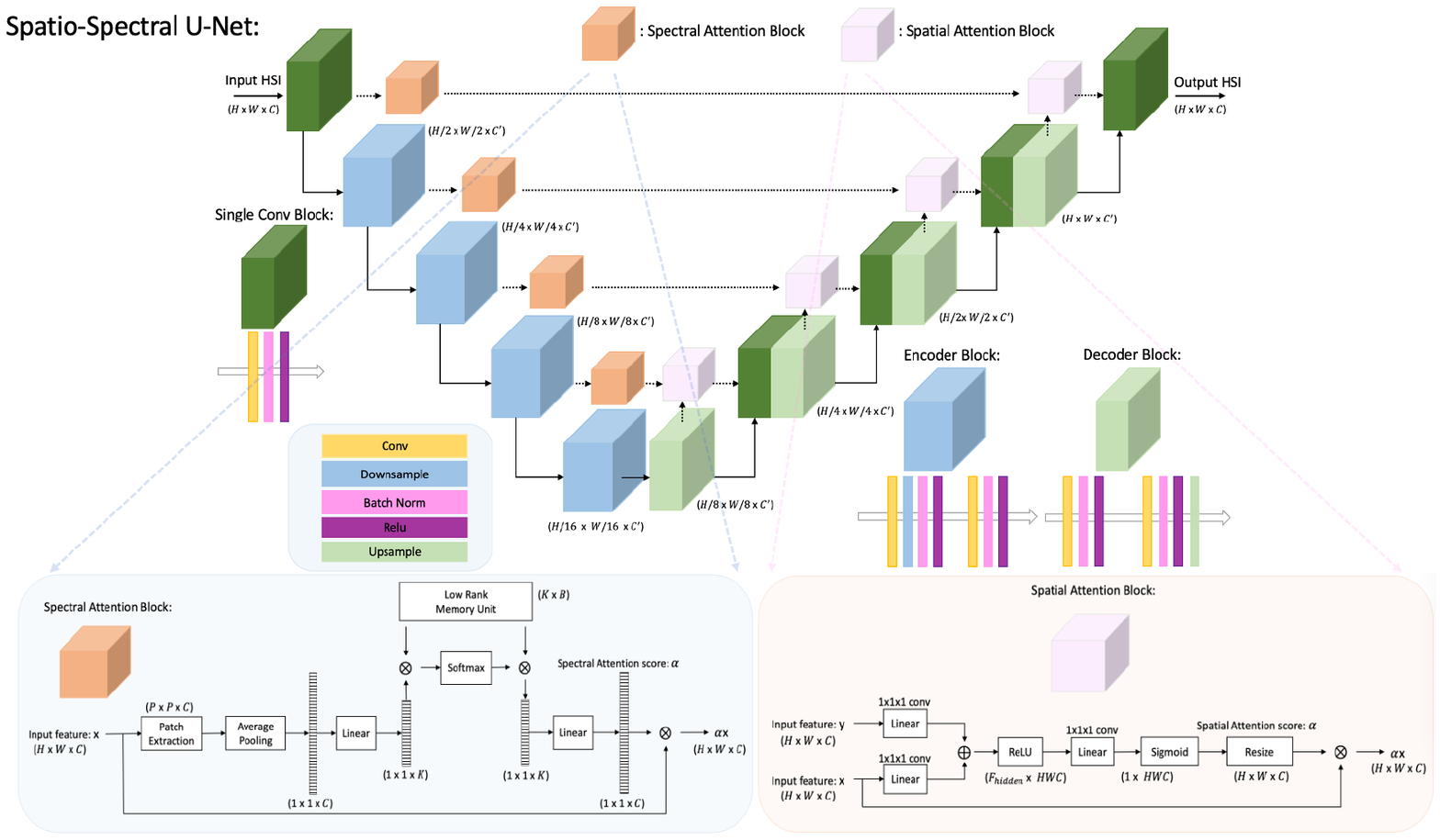}
 }
 \vspace{-1.4cm}
  \caption{Spatio-spectral U-Net used in the experiments.}
  \label{Network_Structure}  
\end{figure*} 
Consistency (MC) Loss and the EI loss. For the HSI inpainting tasks, reconstruction from the MC loss alone is insufficient, as there are possible infinite number of solutions $f(\boldsymbol{y})$ that can fit $\boldsymbol{x} = \mathbf{M} f(\boldsymbol{y})$. \textit{i,e.} The MC loss only guarantees the range-space of $\Tilde{\mathbf{M}}$ being recovered, however, it allows for the null-space to be any trivial values. Hence, we enforced the EI constraint with a weight parameter $\alpha$.

\subsection{Spatio-spectral Attention Blocks}
To reduce the spectral redundancy, a channel attention block \cite{li2023spectral_transformer} is introduced to adaptively assign weights to different channels during training. More specifically, the channel attention block squeezes the input feature maps along the third dimension and uses a low-rank memory unit to learn and focus on the specific spectral bands to inpaint. It has a user-input rank, which resembles the reduction factor proposed in \cite{spectral_attention}. 
For the HSI inpainting task, since a group of pixels in an HS image may represent the same or very similar materials, it is here proposed to combine the channel attention block with spatial attention block \cite{spatial_attention} to capture the correlations between adjacent pixels in the spatial domain as well. The overall structure of the hybrid spatio-spectral attention block is shown in Figure \ref{Network_Structure}. 

\section{Implementation Details}
We evaluate the proposed EI Algorithm on three publicly available Hyperspectral datasets:
\begin{enumerate}
    \item The Chikusei airborne hyperspectral dataset \cite{Chikusei}, the test HS image consists of 192 channels and was cropped to 144 × 144 pixels size.
    \item The Indian Pines dataset from AVIRIS sensor \cite{Indian_Pine}, the test HS image consisted of 200 spectral bands, with spatial size 145x145.
    \item The Botswana dataset acquired by the NASA EO-1 satellite \cite{Botswana_dataset}, the test HS image consists of 145 channels and was cropped to 144 × 144 pixels size.
\end{enumerate}
We simulated different shapes of masks $\rm M$ and applied them to all the spectral bands of the test image. The structure of U-Net and the associated parameters follows \cite{DHP}, where for the spectral attention block, we set the rank $K$ in the low-rank memory unit (see Figure \ref{Network_Structure}) to 4 and keep it fixed across all the test image. \footnote{The default setting of $K$ is 16 in \cite{li2023spectral_transformer, spectral_attention}. However, it was found that setting $K$ = 4 tends to slightly improve the performance. In general, Hyper-EI is quite robust to the choice of $K$.} On the implementation of neural network $f_{\theta}$.
In the implementation of the loss function, we chose shifting as the group transformation $T_{g}$ with a group size $T = 7$, and a weight $\alpha =1$. We use Adam optimizer, and the learning rate is set to 0.01. Two performance metrics: Mean Signal-to-Noise Ratio (MPSNR) and Mean Structural Similarity (MSSIM) are adopted to evaluate the performance in all experiments. 

\begin{table*}
\centering
\begin{tabular}{c rrrrr}
\hline\hline
Methods & Input & DHP \cite{DHP}  &PnP-DIP \cite{pnp_dip}& R-DLRHyIn \cite{R_DLRHyIn} & Hyper-EI (\textcolor{red}{Ours})\\
\hline 
MPSNR $\uparrow$ & 23.753 & 38.551($\pm$ 0.81)  & 39.102($\pm$ 0.31) & 39.437($\pm$ 0.65) & $\boldsymbol{41.584}$($\pm$ 0.34) \\ 
\hline 
MSSIM $\uparrow$ &0.405 & 0.897($\pm$ 0.002)  & 0.905($\pm$ 0.001) & 0.917($\pm$ 0.002) & $\boldsymbol{0.931}$($\pm$ 0.001)  \\
\hline
\end{tabular}
\caption{Performance comparing with other methods. The average and variance over 20 sample generation is shown here} 
\label{Table_comparing}
\vspace{-5mm}
\end{table*} 

\begin{figure*}
 \vspace{-2.5cm}
\centering\includegraphics[width=0.95\textwidth,height=1.27\textwidth,]{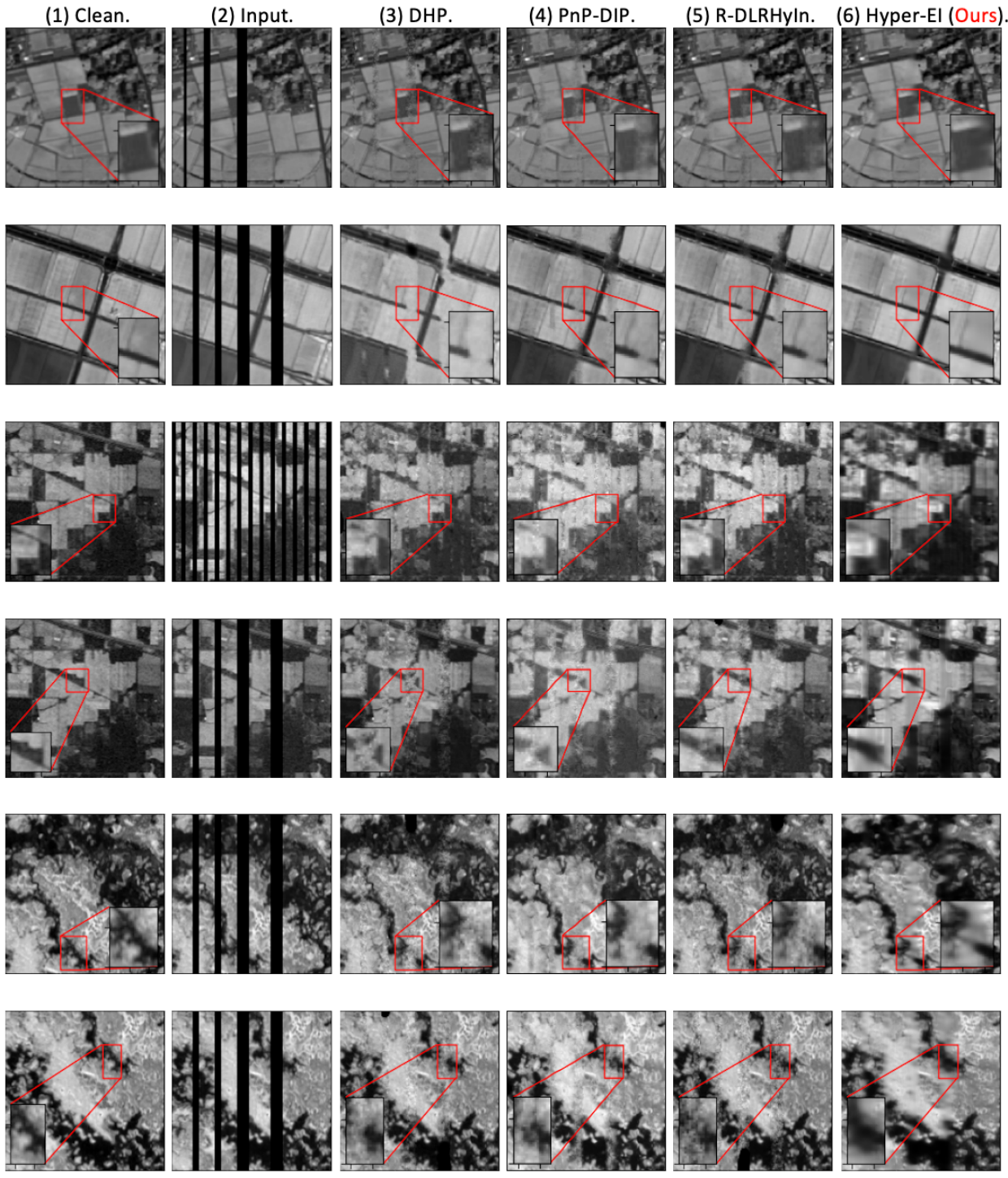}
     \vspace{-2cm}
  \caption{Inpainting performance of Hyper-EI on different HS datasets. Specifically, the top two test samples were selected from the Chikusei dataset, the middle two from the Indian Pine dataset, and the bottom two from the Botswana dataset. From left to right: (1) Clean Image, (2) Input Corrupted Image, (3) DHP, (4) PnP-DIP, (5) R-DLRHyIn, (6) Hyper-EI.}
   \label{inpainting_performance}  
\end{figure*} 

\section{Experimental Results}
The performance of the proposed Hyper-EI algorithm is compared with DHP baseline \cite{DHP}, PnP-DIP \cite{pnp_dip} and R-DLRHyIn \cite{R_DLRHyIn}. For consistency, we used the same Skip-Net in \cite{DHP} as the backbone for the DHP, PnP-DIP and R-DLRHyIn, and fine-tuned the injected regularization noise strength of the input as suggested in \cite{DIP, DHP}. For the proposed Hyper-EI algorithm, we replaced the skip connection with the spatio-spectral blocks of the same complexity to ensure a fair comparison. The numerical results and inpainting performance are reported in Table \ref{Table_comparing} and Figure \ref{inpainting_performance}, respectively. All results are averaged across 20 experiments to account for the impacts of different seeds on the training of DHP/Hyper-EI. \\
R-DLRHyIn and PnP-DIP are two extensions of DHP, the former combines the conventional DHP with a smooth low-rank regularization term to promote the low-rankness of the inpainted HS image, which is the closest method compared to ours in terms of computational complexity. 
From the experimental results in Figure \ref{inpainting_performance}, it was observed that Hyper-EI outperforms the DHP baseline by a significant margin, the inpainted areas of Hyper-EI are more consistent with the background pixels, while other methods have severe distortions at the edges. For example, in the second test sample, the distortion can still be visually identifiable for PnP-DIP and R-DLRHyIn. Besides, Hyper-EI can preserve more textures in the inpainted regions and alleviate the effects of over-smoothing by the denoiser compared to PnP-DIP. 
\begin{table}[H]
\centering
\scalebox{0.92}{
\begin{tabular}{c rrrrrr}
\hline\hline
Methods & Input & No Attention & Spatial & Spectral & Spatio-Spectral\\
\hline 
MPSNR$\uparrow$ & 25.200 & 36.428 & 37.154 & 36.932 & $\boldsymbol{37.665}$\\ 
\hline 
MSSIM $\uparrow$ & 0.8268 & 0.942 & 0.956 & 0.949 & $\boldsymbol{0.970}$\\ 
\hline
\end{tabular}
}
\caption{Ablation test on the spatio-spectral attention blocks} 
\label{ablation_attention}
\end{table} 
\vspace{-0.5cm}
\noindent In Table \ref{ablation_attention}, we provide an ablation test on the inpainting performance of the Hyper-EI algorithm with/without attention blocks, which showed that both spatial attention and spatial attention blocks are important to the success of Hyper-EI. Hyper-EI can be seen as an implicit prior which learns the acquisition physics (in our case, the inpainting masks $\mathbf{M}$) by leveraging the potential subspaces spanned by multiple operators. 

\section{Conclusion}
Hyper-EI is a promising solution for HSI inpainting tasks, which not only generate realistic image pixels in the missing areas, but the images have also smoother contents at the edges. The proposed method exploits spectral and spatial redundancy of HSIs by the introduced spatio-spectral attention blocks and requires no training data except the input (corrupted) image. Extensive experiments on real HS data demonstrate the superiority of the proposed Hyper-EI algorithm over existing self-supervised methods. Although our main focus in this paper is HSI inpainting, we believe that Hyper-EI can easily extend to other HSI inverse problem tasks such as denoising, compressive sampling and super-resolution, which we leave them for future work.

\bibliographystyle{unsrt}

\bibliography{Bibliography} 

\end{document}